\documentclass[letterpaper]{article} 
\usepackage{aaai2026}  
\usepackage{times}  
\usepackage{helvet}  
\usepackage{courier}  
\usepackage[hyphens]{url}  
\usepackage{graphicx} 
\usepackage{natbib}  
\usepackage{caption} 
\usepackage{algorithm}
\usepackage{algorithmic}
\usepackage{booktabs}
\usepackage{multirow}
\usepackage{adjustbox}
\usepackage{makecell}
\usepackage{newfloat}
\usepackage{listings}
\usepackage{dsfont}
\usepackage{amsmath}

\DeclareCaptionStyle{ruled}{labelfont=normalfont,labelsep=colon,strut=off} 
\floatstyle{ruled}
\newfloat{listing}{tb}{lst}{}
\floatname{listing}{Listing}
\title{AAAI Press Formatting Instructions \\for Authors Using \LaTeX{} --- A Guide}
\title{Do Retrieval Augmented Language Models Know When They Don't Know?}
\author {
    Youchao Zhou\textsuperscript{\rm 1,\rm 3}\footnote{This work was done during an internship at SMU},
    Heyan Huang\textsuperscript{\rm 1,\rm 3}\footnote{Corresponding Author},
    Yicheng Liu\textsuperscript{\rm 1},
    Rui Dai\textsuperscript{\rm 1},
    Xinglin Wang\textsuperscript{\rm 1},
    Xingchen Zhang\textsuperscript{\rm 1},
    Shumin Shi\textsuperscript{\rm 1},
    Yang Deng\textsuperscript{\rm 2}
}

\affiliations {
    \textsuperscript{\rm 1}School of Computer Science and Technology, Beijing Institute of Technology\\
    \textsuperscript{\rm 2}School of Computing and Information Systems, Singapore Management University\\
    \textsuperscript{\rm 3}Southeast Academy of Information Technology, Beijing Institute of Technology\\
    \{ yczhou, hhy63, lyc2024, ruidai, wangxinglin, zxc2024, bjssm \}@bit.edu.cn, ydeng@smu.edu.sg
}

\usepackage{bibentry}
\newcommand{\indicator}[1]{\mathds{1}{#1}} 

\begin{document}

\maketitle

\begin{abstract}
Existing large language models (LLMs) occasionally generate plausible yet factually incorrect responses, known as hallucinations. Two main approaches have been proposed to mitigate hallucinations: retrieval-augmented language models (RALMs) and refusal post-training. However, current research predominantly focuses on their individual effectiveness while overlooking the evaluation of the refusal capability of RALMs. Ideally, if RALMs know when they do not know, they should refuse to answer.
In this study, we ask the fundamental question: Do RALMs know when they don’t know? Specifically, we investigate three questions. First, are RALMs well calibrated with respect to different internal and external knowledge states? We examine the influence of various factors. Contrary to expectations, when all retrieved documents are irrelevant, RALMs still tend to refuse questions they could have answered correctly. 
Next, given the model’s pronounced \textbf{over-refusal} behavior, we raise a second question: How does a RALM’s refusal ability align with its calibration quality? Our results show that the over-refusal problem can be mitigated through in-context fine-tuning. However, we observe that improved refusal behavior does not necessarily imply better calibration or higher overall accuracy.
Finally, we ask: Can we combine refusal-aware RALMs with uncertainty-based answer abstention to mitigate over-refusal? We develop a simple yet effective refusal mechanism for refusal-post-trained RALMs that improves their overall answer quality by balancing refusal and correct answers.
Our study provides a more comprehensive understanding of the factors influencing RALM behavior. Meanwhile, we emphasize that uncertainty estimation for RALMs remains an open problem deserving deeper investigation.
\end{abstract}

\begin{links}
    \link{Code}{https://github.com/zuochao912/refusal-ability-of-retrieval-augmented-LLMs}
    \link{Extended version}{https://arxiv.org/abs/2509.01476}
\end{links}

\section{Introduction}

Existing large language models (LLMs) have demonstrated remarkable performance across a wide range of challenging tasks. However, they occasionally generate plausible yet factually incorrect responses—a phenomenon commonly known as hallucinations \cite{2020-nips-rag,2025-tois-halluSurvey}. Prior research has primarily addressed this issue through two approaches: retrieval-augmented generation (RAG) \cite{2020-nips-rag,2023-tacl-inContextRAG} and refusal post-training \cite{2024-naacl-rtuning,2025-aaai-dyRtuning}. RAG leverages external knowledge sources to provide contextual grounding, enabling retrieval-augmented language models (RALMs) to answer queries beyond their internal (parametric) knowledge. In contrast, refusal post-training aims to enhance a model’s ability to proactively abstain from answering when uncertain.

\begin{figure}[t]
  \includegraphics[scale=0.8,width=\linewidth]{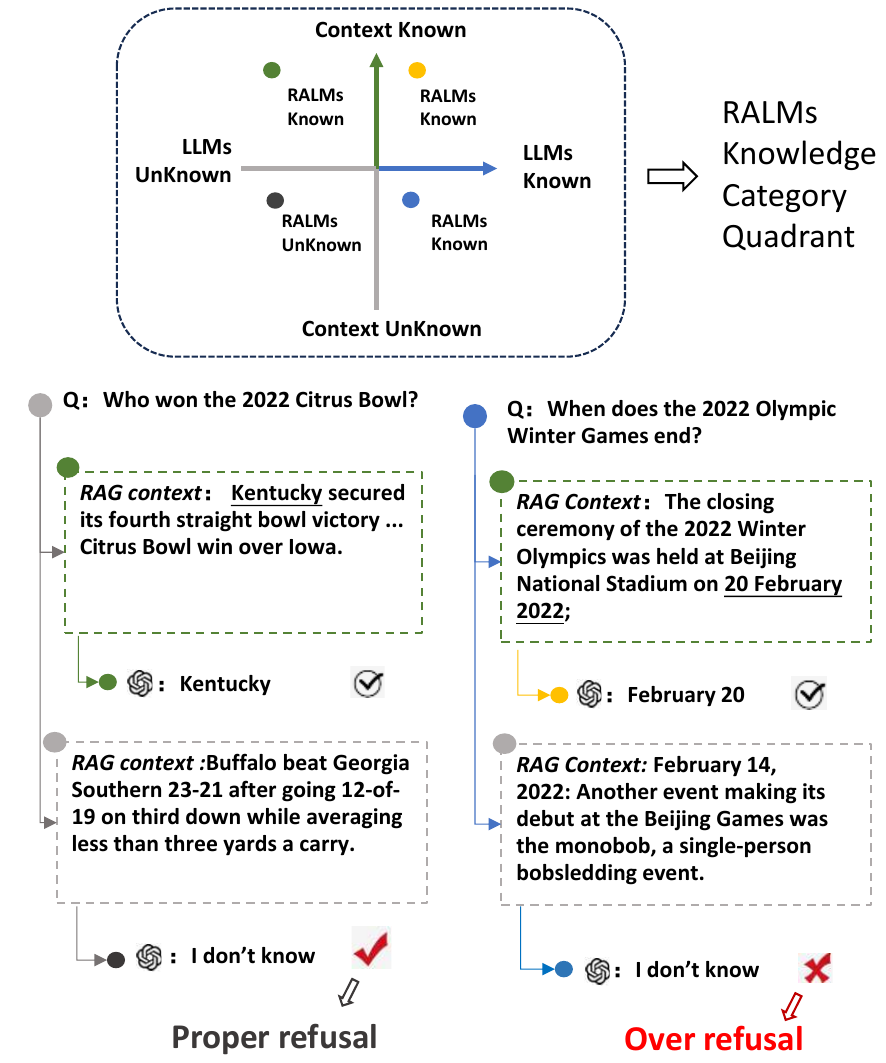}
  \caption{An illustration of the knowledge boundary of a RALM and the corresponding answer correctness. We divide the knowledge state into four quadrants based on the model’s internal knowledge and the knowledge provided by external context. The question at the gray dot lies outside the model’s knowledge boundary, whereas the question at the blue dot lies within it. However, given irrelevant context, the model may still refuse to answer the blue-dot question. }
  \label{fig: LLMkn_example}
\end{figure}

Although both methods are widely adopted, prior work has predominantly emphasized their individual effectiveness while overlooking systematic evaluation of the refusal capabilities of RALMs. Given that LLMs are sensitive to the quality and relevance of retrieval contexts \cite{2024-tacl-robustRalm,2024-sigir-ragnoise}, a refusal-trained model might mishandle unreliable external information and become uncertain even when it internally possesses correct knowledge. As shown in Figure~\ref{fig: LLMkn_example}, RALMs may over-refuse questions that they would otherwise answer correctly when confronted with irrelevant documents. To address this gap, we pose the fundamental question: \textit{Do RALMs know when they do not know?}

Specifically, in this work, we study three critical research questions (RQs). First, \textit{are RALMs well calibrated with respect to different internal and external knowledge states?} (\textbf{RQ1})  Ideally, if RALMs are well calibrated (know when they don't know), they can refuse to answer, or users can post-hoc reject their answers based on model uncertainty. We categorize knowledge states as shown in Figure \ref{fig: LLMkn_example} and quantify the knowledge state of RALMs using uncertainty estimates. We also explicitly consider refusal behavior, which has been overlooked in prior work on uncertainty estimation. While models demonstrate improved calibration when a supportive document exists within otherwise irrelevant contexts, we find that RALMs exhibit significant \textbf{over-refusal} behavior, particularly when confronted with exclusively irrelevant contexts; that is, LLMs still tend to refuse questions they could have answered correctly.

Second, given the over-refusal tendency observed in RALMs, we pose our second research question: \textit{How does a RALM’s refusal ability align with its calibration quality?} (\textbf{RQ2}) We modify the refusal behavior of RALMs using two instruction-tuning-based methods: Refusal-Aware Instruction Tuning (R-tuning)~\cite{2024-naacl-rtuning} and In-Context Fine-Tuning (ICFT)~\cite{2025-arxiv-icft,2025-icml-icftGood}. Our results show that the over-refusal problem is mitigated by ICFT but exacerbated by R-tuning. However, we observe that improved refusal performance does not necessarily imply better calibration or higher answer accuracy. We attribute these discrepancies to changes in robustness and contextual faithfulness.

Lastly, given the difficulty of balancing refusal and response competence based solely on the behavior of LLMs themselves, we investigate our third research question: \textit{Can we combine refusal-aware RALMs with uncertainty-based answer abstention to mitigate over-refusal?} (\textbf{RQ3}) Building on our previous findings, we leverage uncertainty and its variation to infer the knowledge state of RALMs, and then decide whether to answer a question with or without retrieved context, or to abstain altogether.

Our contributions are threefold: 1) We investigate the uncertainty calibration of RALMs and conduct a comprehensive analysis of key factors, including context variation and different knowledge states (internal vs. external knowledge). 2) We identify and characterize the over-refusal problem, and then examine the relationship between refusal behavior and calibration. In particular, we study whether existing refusal tuning exacerbates over-refusal in LLMs and provide further explanations. 3) We design a simple yet effective refusal method for RALMs, informed by the above findings.

\section{Related Works}
\textbf{Knowledge Boundary of LLMs.}
Identifying the knowledge boundary of an LLM helps delineate the limits of its knowledge \cite{sigir25-kb}. This notion is also described as “knowing what you don’t know” \cite{2023-acl(findings)-llmKnowDont,2024-emnlp-sayMore}, which is crucial for assessing the practical applicability of LLMs. \citet{2025-acl-knBoundSurvey} formally categorizes the knowledge boundary with respect to prompt and model sensitivity. However, these works mainly focus on the LLMs’ internal knowledge.
Hallucinations typically occur when users’ requests fall outside the LLM knowledge boundary \cite{2025-tois-halluSurvey}. The primary approach to mitigating hallucinations is retrieval-augmented generation (RAG). RAG \cite{2020-nips-rag} is a convenient approach at inference time, where the retrieved context fills the knowledge gap. More advanced RAG variants leverage LLM self-generated rationales \cite{2024-iclr-instructrag}, perform post-retrieval knowledge selection \cite{2024-iclr-recomp,2024-emnlp(findings)-refiner}, or adopt dynamic retrieval strategies \cite{2024-NAACL-adaptiveRAG}. Recent dynamic RAG methods \cite{2024-iclr-selfrag,2024-acl-dragin} still rely on uncertainty estimates and manually chosen thresholds to decide when retrieval is necessary; even though the system’s knowledge may evolve dynamically, these thresholds remain static. This implicitly assumes that the model is always well calibrated. To the best of our knowledge, no prior work has systematically analyzed the factors that influence the uncertainty of RALMs, and our study fills this gap.

\noindent\textbf{Refusal Method of LLMs.}
Refusal behavior has predominantly been studied at the post-training stage \cite{2025-tacl-refusalSurvey}. Existing work mainly focuses on instruction tuning \cite{2024-naacl-rtuning,2025-aaai-dyRtuning,2024-nips-taughtDontKnow} 
and refusal-alignment training \cite{2024-nips-assistDonKnon,2025-acl-dta}. 
In these setups, instances where the model is uncertain or produces incorrect answers are typically labeled as “should-refuse” examples. Another line of work controls refusal at inference time \cite{2024-acl-abstain}, where uncertainty estimates are used to abstain from answering by thresholds.

\noindent\textbf{Uncertainty Estimation.}
It is crucial for LLMs to recognize their limitations and to express calibrated confidence when responding to users \cite{2023-acl(findings)-llmKnowDont}. Current research typically treats uncertainty and confidence as opposite quantities \cite{2024-tmlr-eigvUe}; that is, the higher the uncertainty of an LLM, the lower its confidence. \citet{2024-naacl-UEsurvey} divide uncertainty estimation (UE) methods for LLMs into white-box and black-box approaches. White-box methods are suitable for open-source LLMs, where internal states are accessible \cite{2023-arxiv-logitUE}. By contrast, black-box methods rely solely on model responses for UE and therefore have broader applicability. Recent work discusses the UE of RALMs \cite{2025-acl-adaptiveRAGUE} and Language Reasoning Models \cite{2025-arxiv-LRMdontknow,2025-arxiv-LRMUQ}. However, these studies do not construct controlled experimental settings to analyze the influence of specific factors, and they neglect the model’s refusal behavior.
\section{Preliminary}
\begin{figure}[t]
  \centering
  \includegraphics[width=0.5\linewidth,keepaspectratio]{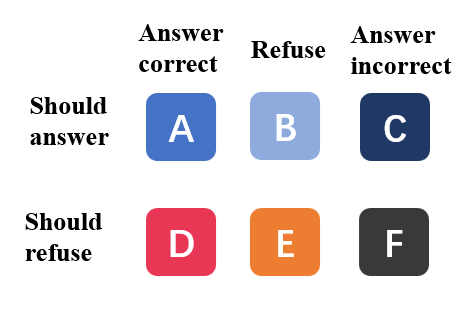}
  \caption{Refusal and answer confusion matrix. “Should answer/refuse” is the ground truth label while “answer correct/incorrect”, refuse is the response situation.}
  \label{fig: prelim}
\end{figure}

We briefly describe the concept of proper refusal and over-refusal. We illustrate the refusal and answer and their correctness situation as in Figure \ref{fig: prelim}. According to \cite{2024-acl-abstain}, the questions could be divided into "should refuse" and "should answer". If LLMs tend to give false answers, which means that LLMs do not entail knowledge, then they should refuse the question. Thus the proper refusal rate is $\frac{E}{D+E+F}$  and the over-refusal rate is $\frac{B}{A+B+C}$. Notice that the “C” and “D” parts exist in our settings. This arises from the threshold used under repeated sampling and the model’s prompt sensitivity. 



\section{Methodology}

\subsection{Uncertainty Estimation Methods}
We primarily adopt black-box UE methods to quantify the confidence of LLM responses, as they are more broadly applicable. Following \cite{2025-acl-adaptiveRAGUE}, we select three categories of well-performing UE methods.

\subsubsection{Verbalization-based UE}
This class of methods leverages the LLM's self-awareness and expressive ability by eliciting explicit confidence estimates for its answers via prompting. We design four different prompts following \cite{2023-emnlp-verbUE}. These prompt variants mainly differ in (i) whether the answer and its uncertainty estimate are produced within the same conversation turn, and (ii) the number of generations elicited. Detailed prompt descriptions are provided in Appendix~A.

\subsubsection{Consistency-based UE}
This class of methods is based on the assumption that more consistent answers indicate higher model confidence. \citet{2025-aaai-LLMcali} propose an alternative approach to quantifying the uncertainty of LLMs and apply it to decoding strategies such as self-consistency. We formalize three types of consistency-based measures as follows.
For a given input $x$ and an LLM $M(\cdot)$, we generate $m$ responses $\{r_1, r_2, \dots, r_m\}$ and decide the final answer via majority voting:
\[
\bar{r} = \arg\max_{r} \sum\nolimits_{i=1}^{m} \indicator{I}(r_i = r),
\]
where $\indicator{I}(\cdot)$ is the indicator function.

The first measurement $Agree(\cdot)$ is based on agreement among answers:
\begin{equation}
    Agree(\bar{r}) = \frac{1}{m} \sum\nolimits^m_{i=1} \indicator({r_i=\bar{r}}),
\end{equation}
where the agreement indicator could be implemented as semantic or lexical agreement, or LLM-as-judge.

The second measurement $Ent(\cdot)$ is entropy-based and rescales the weights of each answer. It is computed as:
\begin{equation}
    Ent(r)=1 - (- \frac{1}{log|\bar{r}|} \sum\nolimits^{|\hat{r}|}_{i=1} p_i log(p_i)),
\end{equation}
where $\hat{r}$ is the set of duplicated answers, $p_i$ is the probability of the unique answer ${r_i}$. 

The final measurement $FSD(\cdot)$ balances the two ways, which is based on the top two most-voted responses $\bar{r}$ and $\bar{\bar{r}}$:
\begin{equation}
FSD(r)=Agree(\bar{r}) -Agree(\bar{\bar{r}})).
\end{equation}

\subsubsection{Similarity Matrix based UE} This kind of methods consider the similarity of all responses. We use two features,including degree and eigenvalue of the similarity matrix following \cite{2024-tmlr-eigvUe}. The formulations are in the Appendix A.

\subsection{Refusal Post-Training Methods}
We aim to adjust the proactive refusal behavior of RALMs. We adopt two refusal instruction tuning (RIFT) methods, namely R-tuning and in-context fine-tuning (ICFT), due to their broad adoption. Further implementation details are provided in Appendix A.

\textbf{R-tuning.}
R-tuning \cite{2024-naacl-rtuning} is a simple yet effective method for teaching LLMs to issue appropriate refusals. Its workflow typically consists of two stages. In the first stage, the questions that the LLM cannot answer correctly are detected. In the second stage, training data are constructed and instruction tuning is performed. For questions outside the model's knowledge boundary, we assign refusal targets such as “I don't know”.

\textbf{In-Context Fine-Tuning.}
\citet{2025-icml-icftGood,2025-arxiv-icft} find that inserting positive context into prompts during instruction tuning improves LLM accuracy. However, they generally append only positive context and train the model to generate correct answers. \citet{2024-ACL-robustICFT,2024-ICLR-robustRALM} adopt a similar strategy but optimize a corresponding training objective to enhance robustness and faithfulness.
In our work, we extend this framework to the refusal setting. For each training example, we insert not only positive context but also negative context. We set the training targets to either a correct answer or a refusal expression according to the knowledge-state quadrant of the RALM, as illustrated in Figure~\ref{fig: LLMkn_example}. When the knowledge is unknown to the RALM, we set the answer to a refusal expression.

\section{Experiments}

\begin{table*}[t]
  \centering
\scalebox{0.8}{
    \begin{tabular}{cc|ccccc|ccccc}
    \toprule
    \multirow{2}[2]{*}{UE type} & \multirow{2}[2]{*}{UE name} & \multicolumn{5}{c|}{$RGB_{en}$}          & \multicolumn{5}{c}{$RGB_{zh}$} \\
          &       & no context & 0p10n & \multicolumn{1}{p{2.835em}}{1p9n} & \multicolumn{1}{p{3.125em}}{5p5n} & \multicolumn{1}{p{3.125em}|}{1p19n} & no context & 0p10n & \multicolumn{1}{p{3.335em}}{1p9n} & \multicolumn{1}{p{2.915em}}{5p5n} & \multicolumn{1}{p{3.04em}}{1p19n} \\
    \midrule
    \multirow{4}[2]{*}{Verbalize} & Verb-1s-1 & 0.445  & \textbf{0.139 } & 0.208  & 0.023  & 0.042  & 0.477  & 0.441  & 0.119  & 0.242  & 0.124  \\
          & Verb-1s-5 & 0.253  & 0.186  & 0.182  & 0.160  & 0.179  & \textbf{0.173 } & 0.170  & 0.182  & 0.170  & 0.198  \\
          & Verb-2s-1 & 0.339  & 0.190  & 0.183  & 0.013  & 0.040  & 0.448  & 0.338  & 0.122  & 0.210  & 0.125  \\
          & Verb-2s-5 & 0.225  & 0.190  & 0.176  & 0.124  & 0.178  & 0.204  & \textbf{0.165 } & 0.412  & 0.240  & 0.442  \\
    \midrule
    \multirow{3}[2]{*}{Consistency} & Ent   & 0.126  & 0.305  & 0.030  & \textbf{0.009 } & 0.033  & 0.253  & 0.256  & 0.093  & 0.148  & 0.082  \\
          & Agree & 0.127  & 0.192  & \textbf{0.026 } & 0.010  & \textbf{0.028 } & 0.250  & 0.261  & \textbf{0.078 } & 0.150  & \textbf{0.075 } \\
          & FSD   & \textbf{0.104 } & 0.162  & 0.041  & 0.014  & 0.048  & 0.201  & 0.182  & 0.083  & \textbf{0.122 } & 0.086  \\
    \midrule
    \multicolumn{1}{c}{\multirow{2}[2]{*}{\makecell{Similarity \\ Matrix-based}}} & Eigv  & 0.202  & 0.232  & 0.289  & 0.271  & 0.260  & 0.247  & 0.282  & 0.299  & 0.271  & 0.284  \\
          & Deg   & 0.200  & 0.229  & 0.292  & 0.275  & 0.262  & 0.236  & 0.277  & 0.297  & 0.268  & 0.283  \\
    \bottomrule
    \end{tabular}%
}
    \caption{The Brier score (lower score indicates better calibration) of different UE methods on different RAG settings and datasets.The “ApBn” means A positive chunks and B negative chunks for RAG context settings.}
    \label{tab:calierr}%
\end{table*}%

\subsection{Experimental Setup}
To focus on the model’s knowledge capacity while minimizing the influence of reasoning, we primarily consider simple factual questions with short answers. These questions typically require only a single evidence document to be answered correctly, for which single-step retrieval is sufficient. Additional details are described in Appendix B.

\subsubsection{RALM Models}
We adopt two prevalent families of open-source LLMs, Qwen and LLaMA. Although modern LLMs are multilingual, We find that Qwen has stronger knowledge in Chinese, whereas LLaMA performs better on English knowledge. To better exploit the knowledge of each model family, we evaluate Qwen\footnote{https://huggingface.co/Qwen} on Chinese datasets and LLaMA\footnote{https://github.com/meta-llama/llama3} on English datasets. \textit{In the main text, we mainly report results for models with approximately 7B parameters.}
For the retrieval component, document chunks and positive ground-truth passages are provided by the original datasets. We perform hybrid search and re-ranking using Milvus\footnote{https://milvus.io} to construct high-quality negative examples, taking both semantic and lexical similarity into account to provide sufficient difficulty. 

\subsubsection{Hyper-Parameters}
The generation temperature is set to 0.5, and the number of sampled generations is set to 16, following \cite{2025-aaai-LLMcali}. Other generation hyper-parameters are kept at the default values for the corresponding LLMs. 

\subsubsection{Datasets}
We explore the RALMs' performance in open-domain QA tasks, using three prevalent fact-oriented single-hop question datasets to evaluate the performance of LLMs, including two RAG datasets, CRUD \cite{2025-tois-crud} and RGB \cite{2024-aaai-rgb}, and an QA dataset, NQ \cite{2019-tacl-nq}. Covering both Chinese and English, the datasets are well-suited for testing Qwen and LLaMA series.
NQ and CRUD are large scale QA/RAG datasets suitable for both training and test. RGB is a dataset particular developed for test, including refusal ability of RALMs. 

\begin{figure}[t]
  \setlength{\belowcaptionskip}{-4pt} 
  \includegraphics[width=\linewidth]{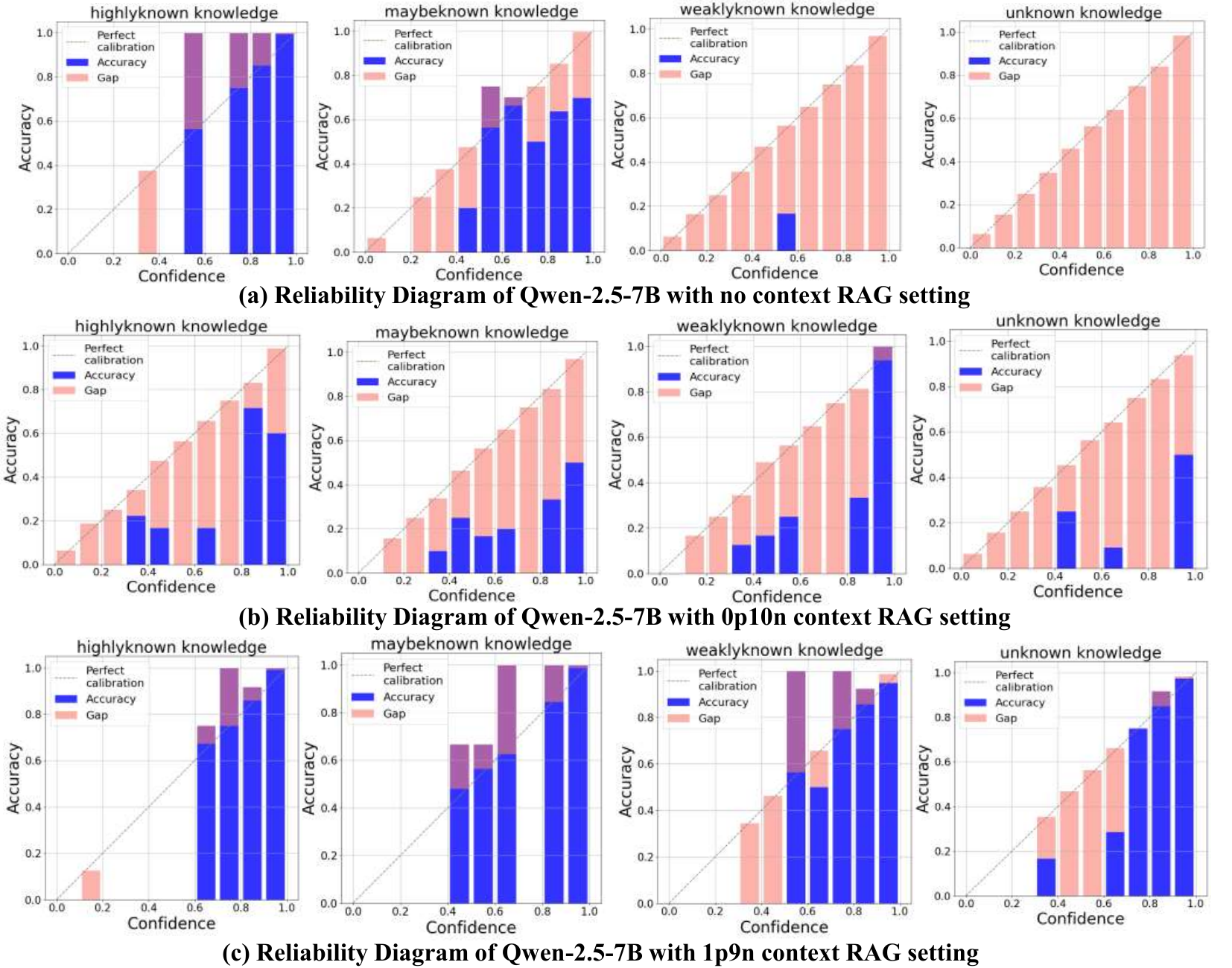}
  \caption{The reliability diagram under different internal and external knowledge states. The blue bar is the precision questions. The pink bar indicates the over-confident gap, and the purple bar indicates the under-confident gap.}
  \label{fig: rq1_reliable_kntype}
\end{figure}


\subsubsection{Answer Judgment}
\label{sec:metrics}
We first assign a knowledge state to each question based on both temperature-sampled and greedy-decoding results, following \cite{2024-emnlp-LlmFtHallu}. This yields four categories: "highlyknown", "maybeknown", "weaklyknown", and "unknown". We treat the former two categories as "should-answer" and the latter two as "should-refuse" according to the precision analysis in Section of RQ1. Following \cite{2025-acl-dta}, we then apply a strict answer-decision workflow to determine whether a model output should be regarded as a refusal or a correct answer, including an LLM-as-a-judge step, exact-match checking, and a refusal-word filter.

\subsubsection{Evaluation Metrics}
Evaluation metrics include accuracy-based and confidence-calibration measures \cite{2024-acl-abstain,2025-acl-dta}. The formal definitions of all metrics are given in Appendix~B, and we briefly summarize them as follows:
\begin{itemize}
    \item \textbf{Accuracy-based metrics}: The answering ability of RALMs is multi-dimensional, reflecting both answer quality and refusal quality.
        \begin{itemize}
            \item Answer Quality (AQ): We report answer precision (Pre), recall (Rec), and F1 for correct answers.
            \item Refusal Quality (RQ): We measure the refusal rate(RR), refusal precision (RPrec), recall (RRec) and F1(RF1).
            \item Overall Quality (OQ):  We report overall accuracy (OAcc), defined as the proportion of outputs that are either correct answers or proper refusals.
        \end{itemize}    
    \item \textbf{Confidence calibration metrics}: We mainly use Brier Score to measure whether the answer confidence measure the answer precision.
\end{itemize}
\subsection{Do RALMs Know When They Don't Know? (RQ1)} 
\label{sec:rq1}
We systematically investigate how prompt variants, positive context position, context quality, and quantity affect the model performance.\footnote{Detailed discussions are in Appendix C} We heuristically varied the numbers of positive and negative examples and examined their impact on the results. In this section, we first examine the calibration error with different UE methods to choose the best one for the following analysis. We then analyze confidence and accuracy in turn as they contribute to the calibration results.

\paragraph{Calibration error of RALMs.}
We exclude refusals for UE, since they are outcome-level decisions co-equal with answering, not comparable to specific answer content. Results are in Table \ref{tab:calierr}.  
The calibration error varies under different RAG settings, and no single method performs best across all scenarios. This aligns with \cite{2025-acl-adaptiveRAGUE}. However, the RALMs become extremely well-calibrated when positive documents exist, especially for verbalize and consistency-based UE methods. This indicates that the UE methods are also acceptable for RALMs. \textbf{As the consistency-based methods perform best generally}, we take their results for further explanation.
We contrast the presence versus the absence of context. We find that when no positive context exists (0p10n), the calibration error becomes worse. And when we insert a single positive context (1p9n), the model becomes extremely calibrated. If we insert more positive context (5p5n), the trend of calibration error vary, become better on $RGB_{en}$ and worser on $RGB_{zh}$. And if we insert more negative context (1p19n), the calibration error does not significantly change. This means that RALMs can sensitively perceive the availability of knowledge. 
\textbf{As we find the key factor is the positive context existence}, the following settings use 10 context chunks as the default.

\begin{figure}[t]
  \includegraphics[width=\linewidth]{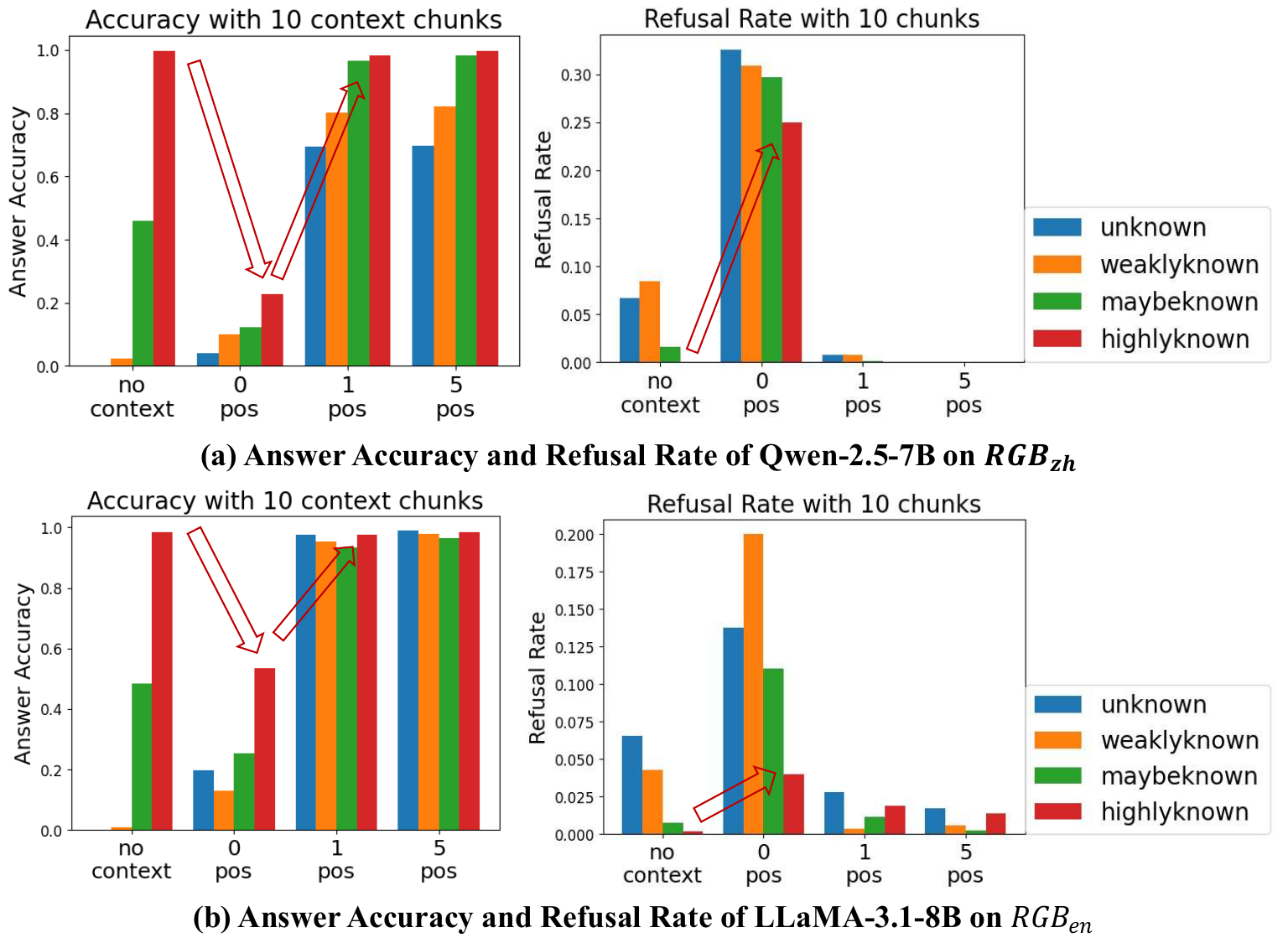}
  \caption{The answer precision (denoted as "accuracy") and refusal rate vary according to the internal/external knowledge states. The whole negative context (0 pos) leads to significant decrease of accuracy and increase of refusal on “highlyknown” questions.}
  \label{fig: rq1_Accs}
\end{figure}

\paragraph{Over-confident or under-confident.}
In this section, we examine how confidence scores vary, given that base LLMs are known to be over-confident \cite{2025-acl-knBoundSurvey} as shown in Figure~\ref{fig: rq1_reliable_kntype}. 
\textit{In the no-context setting}, the “highlyknown’’ type is slightly under-confident, whereas the other types are over-confident. The “highlyknown’’ questions attain relatively high confidence values, while the confidence of the other types is more dispersed.  
However, \textit{in the all-negative-context setting}, the RALMs become strongly over-confident and the confidence scores for all types become highly dispersed. For “highlyknown’’ questions, the LLM could answer correctly without retrieval, yet the observed accuracy is noticeably worse. This indicates that both accuracy and confidence are substantially affected by noisy contexts. Interestingly, “weaklyknown’’ questions achieve higher accuracy under negative contexts, suggesting that the injected noise can have unexpected effects. This finding is consistent with \citet{2024-sigir-ragnoise}, while we further delineate how this effect depends on specific knowledge categories. 
Finally, even \textit{when one positive context is provided}, RALMs tend to be under-confident for most knowledge types, except for the “unknown" category. Across knowledge types, the model attains high accuracy and more concentrated confidence distributions, indicating that RALMs can effectively detect and exploit helpful information. 
In summary, these observations explain the calibration trends in Table~\ref{tab:calierr}: with all-negative context, accuracy generally decreases and confidence becomes more diffuse, whereas with positive context, accuracy improves and confidence becomes more concentrated.

\begin{table*}[t]
  \centering
\scalebox{0.8}{
    \begin{tabular}{cc|ccccccccccc}
    \toprule
    \multicolumn{1}{c}{\multirow{2}[4]{*}{\makecell{RALMs \\ test setting}}} & \multicolumn{1}{c|}{\multirow{2}[4]{*}{\makecell{Method \\ name}}}  & CalErr & OQ    & \multicolumn{4}{c}{AQ}        & \multicolumn{5}{c}{RQ} \\
\cmidrule{3-13}          &       & OaBs $(\downarrow) $  & OAcc$(\uparrow)$   & Pre$(\uparrow)$   & Rec$(\uparrow)$   & F1$(\uparrow)$    & MA$(\downarrow) $   & RR    & OR$(\downarrow) $   & RPre$(\uparrow)$  & RRec$(\uparrow)$  & RF1$(\uparrow)$ \\
    \midrule
          & \multicolumn{1}{c}{} &       &       &       & Qwen-2.5-7B &       &       &       &       &       &       &  \\
    \midrule
    \multirow{6}[2]{*}{no context} & Vanilla & 0.245 & 0.427  & 0.411  & 1.000  & 0.583  & \textbf{0.217 } & 0.027  & 0.000  & \textbf{1.000 } & 0.044  & 0.085  \\
          & R-tuning & 0.191 & 0.457  & 0.395  & 0.857  & 0.541  & 0.336  & \textbf{0.190 } & \textcolor[rgb]{ 1,  0,  0}{\textbf{0.105 }} & 0.719  & \textbf{0.218 } & \textbf{0.335 } \\
          & ICFT (n) & 0.226 & \textbf{0.487 } & \textbf{0.450 } & 0.953  & 0.611  & 0.250  & 0.103  & 0.039  & 0.806  & 0.145  & 0.245  \\
          & ICFT (p) & 0.169 & 0.443  & 0.443  & 1.000  & \textbf{0.614 } & 0.250  & 0.000  & 0.000  & 0.000  & 0.000  & 0.000  \\
          & ICFT (pn) & \textbf{0.167} & 0.440  & 0.440  & 1.000  & 0.611  & 0.243  & 0.000  & 0.000  & 0.000  & 0.000  & 0.000  \\
          & ICFT (w) & 0.181 & 0.423  & 0.414  & 1.000  & 0.585  & 0.296  & 0.017  & 0.000  & 1.000  & 0.028  & 0.055  \\
    \midrule
    \multirow{6}[2]{*}{0p10n} & Vanilla & 0.325 & 0.290  & 0.168  & 0.372  & 0.231  & 0.500  & 0.363  & 0.355  & 0.505  & 0.257  & 0.341  \\
          & R-tuning & 0.408 & 0.457  & 0.294  & 0.195  & 0.235  & 0.184  & \textbf{0.717 } & \textcolor[rgb]{ 1,  0,  0}{\textbf{0.678 }} & 0.521  & \textbf{0.651 } & 0.579  \\
          & ICFT (n) & 0.216 & \textbf{0.620 } & \textbf{0.578 } & 0.709  & \textbf{0.637 } & \textbf{0.158 } & 0.423  & 0.270  & \textbf{0.677 } & 0.541  & \textbf{0.601 } \\
          & ICFT (p) & 0.204 & 0.400  & 0.400  & 1.000  & 0.571  & 0.342  & 0.000  & 0.000  & 0.000  & 0.000  & 0.000  \\
          & ICFT (pn) & \textbf{0.189} & 0.430  & 0.430  & 1.000  & 0.601  & 0.309  & 0.000  & 0.000  & 0.000  & 0.000  & 0.000  \\
          & ICFT (w) & 0.217 & 0.460  & 0.436  & 0.976  & 0.603  & 0.296  & 0.060  & 0.020  & 0.833  & 0.086  & 0.156  \\
    \midrule
    \multirow{6}[2]{*}{1p9n} & Vanilla & 0.079 & \textbf{0.863 } & \textbf{0.863 } & 1.000  & \textbf{0.927 } & \textbf{0.013 } & 0.000  & 0.000  & 0.000  & 0.000  & 0.000  \\
          & R-tuning & 0.127 & 0.830  & 0.853  & 0.960  & 0.903  & 0.033  & 0.070  & 0.066  & 0.524  & 0.212  & 0.301  \\
          & ICFT (n) & 0.164 & 0.787  & 0.835  & 0.881  & 0.858  & 0.033  & \textbf{0.230 } & \textcolor[rgb]{ 1,  0,  0}{\textbf{0.171 }} & \textbf{0.623 } & \textbf{0.531 } & \textbf{0.573 } \\
          & ICFT (p) & \textbf{0.068} & 0.827  & 0.827  & 1.000  & 0.905  & 0.072  & 0.000  & 0.000  & 0.000  & 0.000  & 0.000  \\
          & ICFT (pn) & 0.085 & 0.820  & 0.820  & 1.000  & 0.901  & 0.059  & 0.000  & 0.000  & 0.000  & 0.000  & 0.000  \\
          & ICFT (w) & 0.094 & 0.827  & 0.827  & 1.000  & 0.905  & 0.053  & 0.000  & 0.000  & 0.000  & 0.000  & 0.000  \\
    \bottomrule
    \end{tabular}%
}
  \caption{Evaluation of refusal trained models under different settings. $(\uparrow)$ indicates a higher score is better, and $(\downarrow)$ vice versa. If no arrow is marked, then the score have no directionality. The best result under a RALMs test settings is marked bold and we do not mark those “1.000” scores. The over-refusal score (OR) which is marked in red indicates the worst case.}
  \label{tab:acc_rift}%
\end{table*}%

\paragraph{Precision and refusal rate.}
We begin by analyzing how answer correctness varies. \textit{In the all-negative(0 pos) setting}, we observe a decline on “highlyknown” and “maybeknown” questions and a gain on “weaklyknown” and “unknown” ones compared to the no-context setting. When a positive context exists, the precision significantly increases, especially for unknown and weakly known knowledge. Increasing the count of positives yields no significant gains in precision. This indicates that LLMs are sensitive to both harmful and supportive contexts. While increasing the number of positive and negative examples does not significantly alter the model’s response for fact-oriented questions in this kind of shorter context.
\textit{Then we analyze refusal rate.} In the all-negative (0 pos) setting, we observe an significant increase on all the knowledge types. Considering the LLMs can correctly answer "highlyknown" questions on their own, refusal on those questions are not correct. We identify this phenomenon as \textbf{over-refusal}, which are not observed in previously research. Likewise, the presence of positive chunk markedly reduces refusal. This is consistent with the pattern of accuracy changes.

\paragraph{Summary.}
In this section, we empirically show that RALMs generally “know they don’t know” under no-context and positive-context settings. However, they become over-confident when confronted with negative context and may over-refuse questions whose answers they actually know. 
\subsection{How does RALMs' refusal ability align with its calibration quality? (RQ2)}

\begin{table}[t]
  \centering
  \scalebox{0.9}{
    \begin{tabular}{c|cccc}
    \toprule
    \multirow{2}[4]{*}{Method name} & \multicolumn{2}{c}{DR} & \multicolumn{2}{c}{CU} \\
\cmidrule(r){2-3} \cmidrule(r){4-5}          & no context & 0p10n   & 10p0n & 1p9n \\
    \midrule
    Vanilla & 0.579  & 0.191  & 0.759  & \textbf{0.738 } \\
    R-tuning & 0.444  & 0.138  & 0.750  & 0.682  \\
    ICFT (n) & 0.734  & 0.632  & 0.750  & 0.591  \\
    ICFT (p) & 0.750  & 0.658  & \textbf{0.824 } & 0.723  \\
    ICFT (pn) & \textbf{0.757 } & \textbf{0.691 } & 0.777  & 0.696  \\
    ICFT (w) & 0.704  & 0.684  & 0.770  & 0.703  \\
    \bottomrule
    \end{tabular}%
}
  \caption{Results of denoise rate and positive context utilization.}
  \label{tab:retHand}%
\end{table}%
We adjust refusal ability though the R-tuning and In-context Fine-tuning variants. Considering the knowledge quadrants of Figure \ref{fig: LLMkn_example}, we set four ICFT variants as follows:

\begin{itemize}
    \item ICFT(n) : We append only negative contexts for LLMs, thus the answer of training samples depend on the internal state of LLMs. If internal knowledge entail the question, the answer is original ground truth; else the answer is "I don't known".
    \item ICFT(p) : We append only positive contexts for LLMs. The answers are all set to original ground truth.
    \item ICFT(pn): We append both positive and negative contexts for LLMs and the answers are all set to original ground truth. This is because the LLMs can distinguish the positive context and we want to enhance this ability.
    \item ICFT(w): We include both the ICFT(n) and ICFT(pn) training samples. 
\end{itemize}
We use the training query, only different context and answers to ensure the training fairness. Training and model selection details are in Appendix D \footnote{We also test RL-based refusal-aware methods}.

\paragraph{Response quality of RIFT models}

The response quality of refusal-trained RALMs is multi-dimensional. As shown in Table~\ref{tab:acc_rift}, model performance varies across different RALM settings. 
\textit{In the no-context setting}, ICFT(n) achieves the best overall accuracy (OAcc, OQ), while ICFT(p) performs best in terms of F1 (AQ). The R-tuning model obtains the highest RF1 (RQ), with ICFT(n) ranking second. This may be because the R-tuning training scenario closely matches the test setting, leading to a higher refusal rate (RR) and moderate refusal precision (RPrec). However, the \textbf{over-refusal rate (OR) also increases}, suggesting that R-tuning may harm the model’s self-awareness. The decrease in answer precision (Pre) and the increase in mis-answer rate (MR) support this finding. We will further examine the corresponding change in confidence calibration in the following subsection.
\textit{In the all-negative (0p10n) setting}, ICFT(n) performs substantially better than the other models in terms of OAcc (OQ), F1 (OQ), and RF1 (RQ). Although the over-refusal rate (OR) of R-tuning is the worst, ICFT(n) alleviates this issue and performs better than the vanilla RALMs. Moreover, we find that ICFT variants with positive context substantially reduce over-refusal while maintaining competitive overall accuracy (OAcc, OQ).
Surprisingly, \textit{when positive context is available}, the vanilla RALMs achieve the best OAcc (OQ) and F1 (AQ). From the perspective of RQ, ICFT(n) actually appears to perform the best. However, we emphasize that RQ in this positive-context setting should be interpreted with caution, as we do not relabel the “should-answer’’ set in order to remain consistent with the previous two settings.

\paragraph{Refusal Confidence of RIFT models}
In RQ1 we do not consider the refusal part, we check the overall brier score (OaBs) as in Table \ref{tab:acc_rift}. We notice that the performance of calibration error do not align with overall,answer, or refusal quality. Surprisingly, ICFT with positive context(p/pn) get best calibration performance, though their refusal performance is not good as ICFT(n). This provides support for jointly considering active and passive refusals. We provide confidence distribution illustration Appendix D.



\paragraph{Retrieval handling of RIFT models}
Because a single calibration-error metric cannot fully reflect refusal quality, we introduce retrieval-handling metrics to further explain the results. Intuitively, a model that is more robust to noise is more likely to rely on its internal knowledge. While some methods \cite{2025-ACL-faithfulrag,2025-acl(findings)-contextdpo} explicitly emphasize the context faithfulness of RALMs. We evaluate these abilities using the denoising rate (DR) and the context utilization rate (CU), as reported in Table~\ref{tab:retHand}.
In terms of denoising ability, all ICFT models perform better than the vanilla models, whereas the R-tuning models perform worse than the vanilla baseline. Although the R-tuning methods outperform the vanilla models in OAcc (OQ) and RF1 (RQ), this suggests that R-tuning primarily encourages models to refuse based on their internal states rather than to resist noisy context. However, the R-tuning approach appears to sacrifice the model’s underlying knowledge competence in exchange for a stronger ability to articulate refusals, according to its worse DR performance in no context settings.
In terms of context utilization, we find that ICFT(p) yields better results, while including negative context leads to worse performance in the all-positive (10p) setting. Surprisingly, however, all refusal fine-tuned models perform worse than the vanilla RALMs. This explains why these models perform poorly in scenarios with positive evidence: they tend to refuse internally unknown questions while ignoring the positive context.

\paragraph{Summary}
In this section, our results show that the over-refusal problem is mitigated by In-context fine-tuning, but magnified by R-tuning. The system’s performance should be assessed by jointly considering the model’s confidence, robustness, and context faithfulness. However, we also find that the refusal ability may conflict with the quality of the answer.

\subsection{Mitigating the Over-refusal Issue in RALMs (RQ3)}
Although some refusal-aware RALM models do not support appropriate abstention by themselves, their confidence profiles can still distinguish correct refusals from incorrect ones. To validate whether we can distinguish different knowledge states and enable more appropriate refusals, we first study a simple threshold-based post-refusal technique. Concretely, we follow the thresholds-based refusal at inference stage.

To reduce the negative effects introduced by noisy contexts, we further develop a two-stage refusal technique. In the first stage, we apply a threshold $T_s$ to $U_{\text{LLM}}$ (the uncertainty of the base LLM) to detect whether the answer can be supported by internal knowledge, and a threshold on $\Delta U=U_{\text{RALM}} - U_{\text{LLM}}$ (where $U_{\text{RALM}}$ is the uncertainty of the RALM, which incorporates context) to infer the knowledge state. In the second stage, we apply a refusal threshold in the same way as the baseline, but only when the RALM is classified as “unknown”. All threshold values are selected via grid search on the development set. To better isolate the effect of knowledge on refusal, we compare these methods under an idealized but challenging (0p10n) context configuration. The results are summarized in Table~\ref{tab:rq3_mitigate}. The post-refusal methods achieve higher overall accuracy than their counterparts in Table~\ref{tab:acc_rift}, but they also exhibit a substantially higher over-refusal rate. By first determining the knowledge state of the LLM itself, the model can choose when to rely on its own knowledge, yielding more calibrated confidence estimates and enabling further refusals without overusing harmful negative contexts, especially for ICFT(p) which show better calibration but less tendency to refuse on its own. Finally, we note that \citet{2025-emnlp-infogain} adopts similar information-gain-based method to detect context utility. This further supports our findings, while we provide a more explicit analysis of how knowledge states influence refusal behavior. Additional details for real RAG experiments are provided in Appendix E.

\begin{table}[t]
  \centering
  \scalebox{0.8}{
    \begin{tabular}{cc|ccccc}
    \toprule
    \multicolumn{1}{c}{\multirow{2}[2]{*}{\makecell{Refusal \\ method}}} & \multicolumn{1}{c|}{\multirow{2}[2]{*}{\makecell{Method \\ Name}}} & OQ    & \multicolumn{2}{c}{AQ} & \multicolumn{2}{c}{RQ} \\ \cmidrule(r){3-3} \cmidrule(r){4-5} \cmidrule(r){6-7}
          &       & OAcc  & MA    & AF1   & OR    & RF1 \\
    \midrule
    \multicolumn{7}{c}{0p10n} \\
    \midrule
    \multirow{3}[2]{*}{Post refusal} & Vanilla & 0.437  & 0.145  & 0.167  & 0.770  & 0.570  \\
          & ICFT(n) & 0.673  & \textbf{0.098 } & 0.655  & 0.462  & \textbf{0.690 } \\
          & ICFT(p) & \textbf{0.683 } & 0.240  & \textbf{0.672 } & \textbf{0.243 } & 0.682  \\
    \midrule
    \multirow{3}[2]{*}{Ours} & Vanilla & 0.523  & 0.104  & 0.240  & 0.282  & 0.590  \\
          & ICFT(n) & \textbf{0.729 } & \textbf{0.059 } & \textbf{0.707 } & 0.176  & \textbf{0.731 } \\
          & ICFT(p) & 0.697  & 0.178  & 0.691  & \textbf{0.106 } & 0.698  \\
    \bottomrule
    \end{tabular}%
    }
    \caption{RALMs knowledge state aware refusal technique. }
  \label{tab:rq3_mitigate}%
\end{table}%

\section{Conclusions}

In this work, we investigate whether RALMs “know when they don’t know”. We find that the calibration state of RALMs is greatly influenced by external contexts. In particular, we identify that purely negative contexts severely harm calibration and induce an over-refusal problem. We further study how the refusal quality of RALMs aligns with their calibration and observe that refusal-aware RALMs struggle to handle different RAG settings, due to entangled internal knowledge states and reduced context utilization. Finally, we combine the refusal ability of LLMs with post-refusal methods to balance overall response quality while mitigating over-refusal. Our study offers insights that underscore the need for improved calibration methods and the explicit modeling of dynamically evolving knowledge.

\section{Acknowledgments}
The authors thank all the reviewers for their suggestions and comments. This work is supported by National Natural Science Foundation of China (No.U21B2009). It is also supported by scholarship under the State Scholarship Fund and a visiting to Singapore Management University organized by the China Scholarship Council (CSC). The authors also acknowledge the material support by \textit{Boston Meditech Group} and \textit{Hangzhou Kangyi Health Management Limited Partnership}.

\bibliography{aaai2026}

\end{document}